\documentclass[]{fairmeta}
\usepackage{multirow}
\usepackage{graphicx}   % for \resizebox
\usepackage[utf8]{inputenc} % allow utf-8 input
\usepackage[T1]{fontenc}    % use 8-bit T1 fonts
\usepackage{hyperref}       % hyperlinks
\usepackage{url}            % simple URL typesetting
\usepackage{booktabs}       % professional-quality tables
\usepackage{amsfonts}       % blackboard math symbols
\usepackage{nicefrac}       % compact symbols for 1/2, etc.
\usepackage{microtype}      % microtypography
\usepackage{xcolor}         % colors
\usepackage{amsmath,amsfonts,bm}
\usepackage{algorithm}
\usepackage{algpseudocode}
\usepackage[most]{tcolorbox}
\definecolor{promptbg}{RGB}{245, 245, 245} 
\title{From Solving to Verifying: A Unified Objective for Robust Reasoning in LLMs}

\author[1,2,*]{Xiaoxuan Wang}
\author[1]{Bo Liu}
\author[1]{Song Jiang}
\author[1]{Jingzhou Liu}
\author[1,3,*]{Jingyuan Qi}
\author[1]{Xia Chen}
\author[1]{Baosheng He}

\affiliation[1]{Meta}
\affiliation[2]{UCLA}
\affiliation[3]{Virginia Tech}
\contribution[*]{Work done at Meta}
\abstract{
The reasoning capabilities of large language models (LLMs) have been significantly improved through
reinforcement learning (RL). Nevertheless, LLMs still struggle to consistently verify
their own reasoning traces. This raises the research question of how to enhance the
self-verification ability of LLMs and whether such an ability can further improve reasoning
performance. In this work, we propose \textbf{GRPO-Verif}, an algorithm that jointly optimizes solution generation and
self-verification within a unified loss function, with an adjustable hyperparameter controlling the weight of
the verification signal. Experimental results demonstrate that our method enhances self-verification capability while maintaining comparable performance in reasoning.
}

\date{\today}
\correspondence{Xiaoxuan Wang at \email{xw27@g.ucla.edu}}

\begin{document}

\maketitle
\vspace{-0.1cm}
\section{Introduction}

The reasoning capabilities of large language models (LLMs) have shown remarkable progress in mathematical reasoning\citep{cobbe2021training, zelikman2022star, singh2023beyond,wang2025entropy} and scientific discovery\citep{taylor2022galactica, wang2023scibench}. Reinforcement learning (RL) has played a key role in this advancement, enabling models to generate intermediate reasoning steps and substantially enhance their problem-solving ability\citep{guo2025deepseek}. Despite these successes, LLMs remain highly dependent on external reward signals and continue to struggle with verifying the correctness of their own reasoning\citep{huang2023large, tyen2023llms}.
This limitation raises an important research question: how can we improve the self-verification ability of LLMs, and does enhancing this capability hinder or further strengthen their reasoning performance?

Earlier work used SFT-based methods, training on datasets containing erroneous responses paired with their corrections, but such approaches often introduced distribution shift \citep{chen2023teaching, madaan2023self}. More recent studies explored RL-based strategies for self-verification and self-correction \citep{kumar2024training, jiang2025pag, liu2025trust, zhang2025critique}. While these works demonstrate the potential of reinforcement learning for verification, the effect of explicitly weighting self-verification signals within the training objective has not been systematically studied.

To address this question, we introduce \textbf{GRPO-Verif}, an algorithm that integrates problem-solving and self-verification into a unified objective. During RL training, the model first generates candidate solutions for each problem and then produces verification responses for these solutions. The correctness of the verifications is incorporated into the training signal through an auxiliary loss term, whose influence is controlled by an adjustable weighting parameter. This design enables a systematic study of how self-verification affects both verification accuracy and solution quality.

We evaluate GRPO-Verif using Qwen2.5-3B-Base across four challenging mathematical reasoning benchmarks. Empirical results show that GRPO itself enhances verification accuracy, and GRPO-Verif achieves an additional gain from 32.9\% to 37.1\%, while preserving solution quality at a level comparable to GRPO. This demonstrates that explicit integration of self-verification strengthens verification capability without compromising reasoning ability.
\begin{figure}[h]
    \centering
    \includegraphics[width=0.9\textwidth]{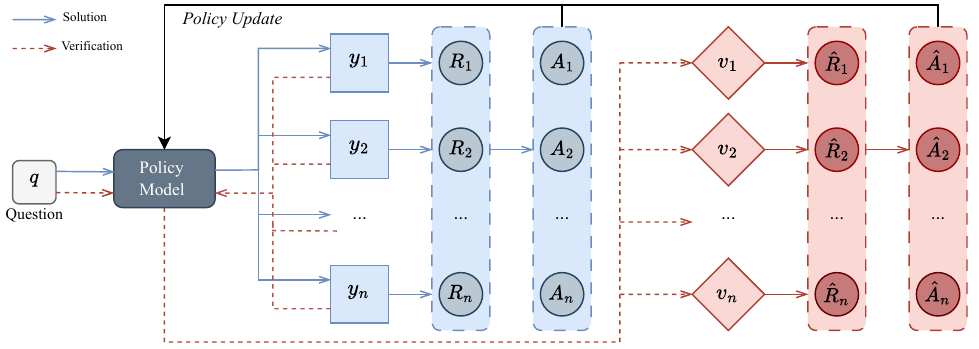} % path to your image file
    \caption{GRPO-Verif Overview. For each question, the policy model generates candidate solutions and computes advantages from solution rewards. Verification responses are then derived from the paired question–solution inputs, with advantages from verification rewards. Both sets of advantages contribute separately to the optimization and are combined into a joint loss to update the policy.}
    \label{fig:sf}
\end{figure}

\section{Method}
\subsection{Preliminary: GRPO}
GRPO is an online Reinforcement Learning algorithm for fine-tuning LLMs \citep{guo2025deepseek}. In the context of LLM policy optimization, let the model policy be parameterized by $\theta$. For each question $q$ in a given set $Q$, a group of solutions $\{y^{(i)}\}_{i=1}^n$ is sampled from the old policy $\pi_{\text{old}}$. We define the reward signals $\{R^{(i)}\}_{i=1}^n$ for the solutions, where each reward indicates the accuracy of the corresponding final answer. The GRPO training objective is formulated as:
\begin{equation}
\begin{aligned}
J_{\text{GRPO}}(\theta) &= 
\mathbb{E}_{q \sim Q, \{y^{(i)}\}_{i=1}^n \sim \pi_{\text{old}}(\cdot|q)} \\
&\quad  \frac{1}{n} \sum_{i=1}^n \frac{1}{|y^{(i)}|} \sum_{t=1}^{|y^{(i)}|}
\min \Big( r^{(i)}_t(\theta)A^{(i)}_t,\; \text{clip}(r^{(i)}_t(\theta), 1-\epsilon, 1+\epsilon)A^{(i)}_t \Big) 
- \beta D_{\text{KL}}[\pi_\theta \| \pi_{\text{ref}}],
\end{aligned}
\label{eq:grpo_obj}
\end{equation}

\begin{equation}
r^{(i)}_t(\theta) = 
\frac{\pi_\theta\!\left(y^{(i)}_t \mid q, y^{(i)}_{<t}\right)}
{\pi_{\text{old}}\!\left(y^{(i)}_t \mid q, y^{(i)}_{<t}\right)}, 
\quad A^{(i)}_t =
\frac{R^{(i)} - \text{mean}\!\left(\{R^{(1)}, \ldots, R^{(n)}\}\right)}
{\text{std}\!\left(\{R^{(1)}, \ldots, R^{(n)}\}\right)}
\label{eq:ratio}
\end{equation}

Here, $\epsilon$ and $\beta$ are hyperparameters that respectively control the clipping range of the probability ratio and the strength of the KL divergence penalty. Recent work has shown that setting $\beta=0$ and using a large clipping threshold can improve performance while reducing computational cost \citep{yu2025dapo}. Accordingly, we adopt this strategy in our implementation. The advantage $A^{(i)}_t$ is computed by normalizing the rewards $\{R^{(i)}\}_{i=1}^n$ with respect to their group mean and standard deviation. 

\subsection{Self-Verification GRPO: GRPO-Verif}
To jointly enhance both solution generation and verification capabilities of large language models (LLMs), we propose \textbf{GRPO-Verif}, an extension of GRPO that incorporates self-verification as an auxiliary objective during training. 

For a given prompt $q$, we sample $n$ candidate solutions $\{y^{(i)}\}_{i=1}^n \sim \pi_{\text{old}}(\cdot|q)$. Each solution is evaluated by a rule-based reward function that measures the correctness of its final answer, yielding scalar rewards $\{R^{(i)}\}_{i=1}^n$. In addition, for each solution $y^{(i)}$, we generate a verification response $v^{(i)}$ using a fixed verification template conditioned on both the question and the solution. These verifications provide an auxiliary supervision signal that encourages the model to produce outputs that are not only correct but also self-consistent. The overall GRPO-Verif objective integrates both solution-level rewards and verification-based signals into a single optimization objective:

\begin{equation}
\label{eq:loss}
\begin{aligned}
J_{\text{GRPO-Verif}}(\theta) &= \mathbb{E}_{q \sim Q, \{y^{(i)}\}_{i=1}^n \sim \pi_{\text{old}}(\cdot|q), \{v^{(i)}\}_{i=1}^n \sim \pi_{\text{old}}(\cdot|q,y^{(i)})} \\
&\quad \frac{1}{n} \sum_{i=1}^n \left( \frac{1}{|y^{(i)}|}\sum_{t=1}^{|y^{(i)}|} r_t^{(i)}(\theta) A_t^{(i)} 
+ \alpha * \frac{1}{|v^{(i)}|}\sum_{t=1}^{|v^{(i)}|} \hat{r}^{(i)}_t(\theta) \hat{A}^{(i)}_t \right),
\end{aligned}
\end{equation}

where the token-level probability ratios for solution $r^{(i)}_t(\theta)$ and verification $\hat{r}^{(i)}_t(\theta)$ are defined as

\begin{equation}
\label{eq:ratio}
r^{(i)}_t(\theta) = \frac{\pi_\theta \big( y^{(i)}_t \mid q, y^{(i)}_{<t} \big)}
{\pi_{\text{old}} \big( y^{(i)}_t \mid q, y^{(i)}_{<t} \big)}, \quad 
\hat{r}^{(i)}_t(\theta) = \frac{\pi_\theta \big( v^{(i)}_t \mid q, y^{(i)}, v^{(i)}_{<t} \big)}
{\pi_{\text{old}} \big( v^{(i)}_t \mid q, y^{(i)}, v^{(i)}_{<t} \big)}.
\end{equation}

An adjustable hyperparameter $\alpha$ controls the relative strength of the self-verification term in the overall loss. 

The reward for a solution $R^{(i)}$ is $+1$ if correct, $-1$ if incorrect, and $-1.5$ if no valid answer is produced. The reward for a verification $\hat{R}^{(i)}$ follows the same scheme: $+1$ for a correct verification, $-1$ for an incorrect verification, and $-1.5$ when no valid judgment is provided. Verifications are generated for the same question but based on different solutions. Because these verification responses are produced independently and without conditioning on one another, their raw rewards are normalized with respect to the mean and standard deviation across all verifications. This normalization stabilizes training and provides a loose estimator for the advantage values. The advantages of solution $A_t^{(i)}$ and verification $\hat{A}^{(i)}_t$ are therefore defined as

\begin{equation}
\label{eq:adv}
A_t^{(i)} = \frac{R^{(i)} - \text{mean}\big(\{R^{(j)}\}_{j=1}^n \big)}
{\text{std}\big(\{R^{(j)}\}_{j=1}^n \big)}, 
\quad 
\hat{A}^{(i)}_t = \frac{\hat{R}^{(i)} - \text{mean}\big(\{\hat{R}^{(j)}\}_{j=1}^n \big)}
{\text{std}\big(\{\hat{R}^{(j)}\}_{j=1}^n \big)}.
\end{equation}

By jointly optimizing for both solution accuracy and verification reliability, \textbf{GRPO-Verif} encourages the model to generate responses that are not only correct but also verifiable, leading to more consistent and trustworthy reasoning behavior.
\begin{algorithm}[ht]
\caption{Self-Verification GRPO (GRPO-Verif)}
\label{alg:grpo-sf}
\begin{algorithmic}[1]
\Require Policy $\pi_\theta$, dataset $Q$, number of generations $n$, clipping threshold $\epsilon$, self-verification weight $\alpha$, learning rate $\eta$, number of epochs $T$
\Ensure Trained parameters $\hat{\theta}$
\State Initialize behavior policy $\pi_{\text{old}} \gets \pi_\theta$

\For{epoch $=1$ to $T$}
  \State Sample a minibatch $\mathcal{B} \subset Q$
  \For{each question $q \in \mathcal{B}$}
    \State Generate $n$ candidate solutions $\{y^{(i)}\}_{i=1}^n \sim \pi_{\text{old}}(\cdot \mid q)$
    \State Assign solution rewards $\{R^{(i)}\}_{i=1}^n$ based on final answers
    \For{each solution $y^{(i)}$}
      \State Generate candidate verification $v^{(i)} \sim \pi_{\text{old}}(\cdot \mid q, y^{(i)})$
      \State Assign verification reward $\hat{R}^{(i)}$ based on verification result
    \EndFor
    \State Compute within-group advantage for each $i$: $A^{(i)}_t$ and $\hat{A}^{(i)}_t$ using Eq.~\ref{eq:adv}
    \State Compute token-level ratio for each $i$: $r^{(i)}_t(\theta)$ and $\hat{r}^{(i)}_t(\theta)$ using Eq.~\ref{eq:ratio}
  \EndFor
  \State Compute objective $J_{\text{GRPO-Verif}}(\theta)$ on $\mathcal{B}$ using Eq. \ref{eq:loss}
  \State Update parameters $\theta \gets \theta + \eta \nabla_\theta J_{\text{GRPO-Verif}}(\theta)$
  \State Refresh behavior policy $\pi_{\text{old}} \gets \pi_\theta$
\EndFor
\State \Return $\hat{\theta} \gets \theta$
\end{algorithmic}
\end{algorithm}
\vspace{-2cm}
\section{Experiment}
\vspace{-0.5em}
\subsection{Experiment Setup}
\textbf{Datasets.}
We randomly select 6k problems from DAPO-Math-17k \citep{yu2025dapo} as training data. We evaluate our method on the following benchmarks: AMC23 \citep{Li2024Numinamath}, MATH \citep{hendrycks2020measuring}, Minerva\_Math \citep{lewkowycz2022solving}, and OlympiadBench \citep{he2024olympiadbench}.

% Describing the model
\textbf{Implementation Details. }
We employ Qwen2.5-3B as our base model \citep{qwen25}. We set the learning rate to $1 \times 10^{-6}$, the batch size to 32, $\beta=0$, the clip rate to 0.28, the number of generations $n$ to 8, the maximum completion length to 2048, and the temperature to 1 for sampling. The coefficient weight $\alpha$ is set to 0.2. Additional experimental details are provided in Appendix~\ref{ap:exp}.
\vspace{-0.2em}
\subsection{Experiment Result}
\vspace{-0.1cm}
\begin{table*}[h]
\centering
\resizebox{0.8\textwidth}{!}{
\begin{tabular}{lcccccc}
\toprule
Model & Setting & AMC23 & MATH & Minerva & OlympiadBench & AVG \\
\midrule
\multirow{3}{*}{Solution} 
 & \textit{default}   & 27.5 & 52.7 & 11.0    & 21.0 & 28.1 \\
 & GRPO               & 45.0 & 63.4 & 19.5  & 25.8 & 38.4 \\
 & GRPO-Verif         & 45.0 & 63.2 & 18.8  & 26.8 & 38.5 \\
\midrule
\multirow{3}{*}{Verification} 
 & \textit{default}   & 12.5 & 26.4 & 10.3    & 15.6 & 16.2 \\
 & GRPO               & 25.0 & 57.7 & 21.0  & 27.7 & 32.9 \\
 & GRPO-Verif         & 40.0 & 60.7 & 19.9  & 27.8 & 37.1 \\
\bottomrule
\end{tabular}
}
\caption{Experimental results in terms of accuracy (\%) on benchmarks for Qwen2.5-3B.}
\label{table:final}
\end{table*}

Table~\ref{table:final} reports the accuracy of GRPO-Verif compared with vanilla GRPO and the base model (\textit{default}) under problem-solving (solution) and self-verification (verification) tasks. As shown, GRPO improves both solution and verification accuracy over the default setting, with gains from 28.1\% to 38.4\% for solutions and from 16.2\% to 32.9\% for verification. GRPO-Verif further raises verification accuracy to 37.1\% while maintaining comparable solution accuracy at 38.5\%. These results demonstrate that reinforcement learning enhances both problem-solving and self-verification, and that GRPO-Verif yields additional improvement in verification without compromising solution quality.

% \subsection{Ablation Study on $\alpha$}

% \newpage
\vspace{-0.2cm}
\section{Related Work}
The focus of traditional verifiers has mainly been on distinguishing responses as either preferred or dispreferred \citep{cobbe2021training, lightman2023let}. More recently, LLM-based approaches have introduced chain-of-thought (CoT) verification, where models generate both an answer and an accompanying reasoning trace \citep{mahan2024generative, zhang2024generative}, with some work further employing reinforcement learning and inference-time scaling to enhance verification capability \citep{chen2025rm, liu2025inference, shi2025heimdall, xu2025unified}. However, these efforts generally treat problem-solving and verification as separate tasks. Recent studies have begun to leverage verification itself as a means of self-improvement \citep{xiong2025self, zhang2025critique, jiang2025pag}. In parallel, RISE \citep{liu2025trust} incorporated self-verification directly into the PPO training loop by combining generation and verification trajectories within a unified batch and optimizing them jointly with a shared critic. By contrast, our work introduces self-verification into GRPO, a critic-free algorithm that employs group-normalized advantages and allows the verification loss to be explicitly weighted as an auxiliary objective.
\vspace{-0.2cm}
\section{Conclusion}
We present \textbf{GRPO-Verif}, a reinforcement learning algorithm that integrates self-verification into GRPO through an auxiliary loss with adjustable weighting. Our experiments show that GRPO improves self-verification accuracy, and GRPO-Verif provides an additional boost while preserving solution quality. These results demonstrate that explicit self-verification can enhance the reliability of large language models without compromising their reasoning ability.

\clearpage
\newpage
\bibliographystyle{plainnat}
\bibliography{main}

% \clearpage
% \newpage
% \beginappendix

\clearpage
\appendix
\section{Implementation Detail}
\label{ap:exp}
\subsection{Reproducibility}
All models are trained on NVIDIA H200 GPUs. The training pipeline is built upon the TRL repository \citep{vonwerra2022trl}. Evaluations are conducted on NVIDIA RTX A6000 GPUs, following the evaluation pipeline of \citet{yang2024qwen2}, with the temperature fixed at 0 to ensure reproducibility.
\subsection{Prompt}
\begin{tcolorbox}[
  colback=promptbg,         % body background
  colframe=black!40,        % border color
  boxrule=0.6pt,            % border thickness
  arc=2mm,                  % rounded corners
  % colbacktitle=prompttitle, % title background color
  coltitle=black,           % title text color
  fonttitle=\bfseries,
  title=Solution Prompt,
  enhanced,
  breakable
]
\begin{verbatim}
<|im_start|>system 
Please reason step by step, and put your final answer
within \\boxed{{}}.<|im_end|>
<|im_start|>user {prompt} <|im_end|>
<|im_start|>assistant
\end{verbatim}
\end{tcolorbox}

% \begin{tcolorbox}[
%   colback=promptbg,         % body background
%   colframe=black!40,        % border color
%   boxrule=0.6pt,            % border thickness
%   arc=2mm,                  % rounded corners
%   % colbacktitle=prompttitle, % title background color
%   coltitle=black,           % title text color
%   fonttitle=\bfseries,
%   title=Solution Prompt,
%   enhanced,
%   breakable
% ]
% \begin{verbatim}
% Given the problem: ```{prompt}```

% Here is a solution: ```{solution}```

% Please verify if this solution is correct. Conclude with:
% \\boxed{PASS} — if the solution is correct
% \\boxed{FAIL} — if the solution is incorrect
% \end{verbatim}
% \end{tcolorbox}

\begin{tcolorbox}[
  colback=promptbg,         % body background
  colframe=black!40,        % border color
  boxrule=0.6pt,            % border thickness
  arc=2mm,                  % rounded corners
  coltitle=black,           % title text color
  fonttitle=\bfseries,
  title=Verification Prompt,
  enhanced,
  breakable
]
Given the problem: \texttt{\`{}\`{}\`{}\{prompt\}\`{}\`{}\`{}}

Here is a solution: \texttt{\`{}\`{}\`{}\{solution\}\`{}\`{}\`{}}

Please verify if this solution is correct. Conclude with:

\verb|\\boxed{PASS}| — if the solution is correct

\verb|\\boxed{FAIL}| — if the solution is incorrect
\end{tcolorbox}

\section{Limitation}
A current limitation of GRPO-Verif is the added computational overhead introduced by generating and training on verification responses. This increases training cost and latency compared with standard GRPO. Future work could address this issue by developing more efficient verification strategies or by leveraging lightweight verification modules that reduce resource consumption while retaining performance gains.

\end{document}